\pdfoutput=1
\documentclass[11pt,UTF8]{article}
\usepackage[]{acl}
\usepackage{times}
\usepackage{latexsym}
\usepackage{CJKutf8}
\usepackage{microtype}
\usepackage{multirow}
\usepackage{graphicx}
\usepackage{booktabs}
\usepackage{textcomp}
\usepackage{amsmath}
\usepackage{xcolor,soul}
\graphicspath{ {./} }

\newcommand{\sentinel}[1]{\texttt{\textlangle{}{#1}\textrangle{}}}
\title{Vec2Gloss: definition modeling leveraging contextualized vectors with Wordnet gloss}

\author{Yu-Hsiang Tseng, Mao-Chang Ku, Wei-Ling Chen, Yu-Lin Chang, Shu-Kai Hsieh \\
Graduate Institute of Linguistics, National Taiwan University \\
\texttt{seantyh@gmail.com},
\texttt{d08142002@ntu.edu.tw},
\texttt{d10142007@ntu.edu.tw}, \\
\texttt{cylchang37@gmail.com},
\texttt{shukaihsieh@ntu.edu.tw} }

\begin{document}
\begin{CJK*}{UTF8}{bsmi}
\maketitle
\begin{abstract}

Contextualized embeddings are proven to be powerful tools in multiple NLP tasks. Nonetheless, challenges regarding their interpretability and capability to represent lexical semantics still remain. In this paper, we propose that the task of definition modeling, which aims to generate the human-readable definition of the word, provides a route to evaluate or understand the high dimensional semantic vectors. We propose a `Vec2Gloss' model, which produces the gloss from the target word's contextualized embeddings. The generated glosses of this study are made possible by the systematic gloss patterns provided by Chinese Wordnet. We devise two dependency indices to measure the semantic and contextual dependency, which are used to analyze the generated texts in gloss and token levels. Our results indicate that the proposed `Vec2Gloss' model opens a new perspective to the lexical-semantic applications of contextualized embeddings. 
\end{abstract}

\section{Introduction}

The rapid advancement of distributed semantic models has achieved remarkable results, with machine performance in some language-related benchmarks either matching or even surpassing that of human non-experts 
\citep{Maru2022,Chowdhery2022}. These successes are often attributed to the complex pretrained language models \citep{Peters2018,Devlin2019,Radford2019,Raffel2020} which are broadly referred to as sentence encodings in the literature \citep{Pavlick2022}. In contrast to the traditional distributional semantic models \citep{Lenci2018,Boleda2020}, the sentence encodings are trained top-down, where processing sentences are the primary goal, and the word-level semantics are emergent properties \citep{Pavlick2022}.

Studies have shown sentence encodings do capture lexical semantics. While the contextualized embeddings of each token are highly intertwined with sentiment and syntax \citep{Yenicelik2020}, one could still access a wealth of information on word-level lexical semantics by averaging the vectors across context and model layers. When configured properly, these emerged lexical representations outperform the explicitly trained static word vector models \citep{Vulic2020}. Arguably, these contextualized embeddings are possibly \emph{sense-aware}. One could build the sense embeddings with which the word sense disambiguation is framed as finding the nearest neighbor of the target word in the sense embedding space~\citep{Scarlini2020}. These studies demonstrated that while sentence encodings are not explicitly trained for word-level semantics, they capture the nuances of word usage to a certain degree.

Nonetheless, challenges regarding their interpretability and capability to represent lexical semantics still remain. There have been many evaluations proposed. One unique approach is definition modeling, which aims to generate a definition given the word. The approach is argued to be a more transparent and direct evaluation of the words' semantic representations \citep{noraset2017definition,gardner2022definition}. In light of  distributional semantic models, the definition modeling could be understood as first encoding the semantic representations into one or a set of vectors, based on which a language model generates the corresponding definitions. Past studies have provided abundant model architecture choices with fruitful results. However, the unique merit of the definition modeling is that one can now analyze the embeddings in a natural language form, i.e., the definitions. Instead of indirectly examining a high dimensional vector through word analogies and similarities, we can again probe into the (distributional) lexical semantics transparently with human's natural language.

The follow-up challenge remains as to how to systematically study the generated definitions, especially when it is produced by a model that may or may not capture the nuances of definition language. In this paper, we investigate the model-generated definitions using a relatively standardized gloss language to train a definition generation model. The gloss dataset comes from the Chinese Wordnet (CWN) ~\citep{Huang2010cwn}\footnote{The data are accessible at \url{https://lopentu.github.io/CwnWeb/}}. Chinese Wordnet differentiates the lexical senses of each word and describes them with a relatively constrained set of glossing rules. We formulate the definition modeling as a vector-to-text task. Specifically, inspired by the sense embedding and the sequence-to-sequence architecture of definition modeling \citep{Scarlini2020,Mickus2019}, we further encode the context-sensitive word sense into an encoding vector, from which the model learns to decode the gloss sentences. We use human ratings to evaluate the generated definitions and propose two indices to examine the contextual and semantic dependencies closely. With these two indices, we conduct gloss and token-level analyses of generated definitions and show that the generated definitions fairly reflect the aspects of lexical semantics.

The overarching goal of this work aims to explore the possibility of gloss generation with only one contextualized vector. We propose that a generation model can be trained on relatively constrained gloss patterns from the fine-grained Wordnet gloss. The performance of the model is evaluated with human raters with the in-depth analysis of generated gloss patterns.\footnote{The code and the rating material are available at the anonymized repository: \url{https://anonymous.4open.science/r/vec4gloss-F2C8/}}

\section{Related Work}

\subsection{Patterns in gloss languages}

Dictionary definitions, or word glosses, are ``language about language'', or ``metalanguage'' \citep{sinclair-1991-corpus,johnson_johnson_1998,hanks-2013-lexical-analysis}. One of the popular metalanguage theories is the Natural Semantic Metalanguage (NSM) \citep{wierzbicka_1972}, which proposes that universal semantic primitives can account for meanings of words. For example, \citet{durst_2004} identifies the features of these primitives as: ``indefinability", ``indespensability", ``universality", and ``combinability". On the other hand, \citet{barque&polguere_2004} have identified that sense descriptions can be categorized into ``word paraphrases" and ``word interpretations" according to their formal nature. (cf. \citealp{pottier_1974} and \citealp{pustejovsky_1998})

While previous studies of metalanguage more frequently adopt a logical or formal semantic approach, the Corpus Pattern Analysis (CPA) proposed by \citet{hanks-2004-cpa} provides us with a new direction into analyzing word glosses from perspectives of syntagmatic patterns. According to \citet{firth1957synopsis}, the meanings of a word are contributed to the context formed by surrounding terms. In the same vein, \citet{hanks-2004-cpa} analyzes concordance lines from corpus to generalize typical patterns of certain words. These groups of words constitute \emph{lexical set} which is united by a \emph{semantic type}. For example, \textit{guns}, \textit{rifles}, \textit{pistols} are a lexical set related to the verb \textit{fire}, and they are under the semantic type of firearms \citep{hanks-2013-lexical-analysis}. 

While not closely following the methodology in CPA, the gloss language in Chinese Wordnet tries to incorporate the lexical sets and semantic types into its gloss. For example, one of the gloss patterns\footnote{See the manual of CWN (in Chinese),  \url{https://lope.linguistics.ntu.edu.tw/cwn/documentation}} for adverbial senses is in the form of `表...的程度', such as `表(超過平常)的程度' (a sense of  \textit{很 hěn} `very'), which literally translates to `describing (exceeding normal) extent.' The same glossing guidelines are created across lexical categories. Therefore, the CWN glosses provide us with a fertile ground to systematically model its gloss language. However, as the gloss patterns are too complex for logical or formal analyses, definition modeling with deep learning is beneficial for exploring the hidden information under these gloss patterns.

\subsection{Definition Modeling}

Definition modeling generates a definition for a target word~\citep{gardner2022definition,noraset2017definition}. \citet{noraset2017definition} leverage hypernym embeddings to generate dictionary definitions. Since it is inevitably difficult to capture the sense of a polysemy given a single word input, \citet{gadetsky2018conditional} incorporate the context words' embeddings and an attention-based skip-gram model to build definition modeling on polysemous words. Recent definition modeling further incorporates other architectures to capture the semantic vectors and definition generation better. To obtain the semantic representations of the target word, models in past studies use recurrent neural networks, variational generative models, and other pretrained language models \citep{ishiwatari-etal-2019-learning,Reid2020,Zhang2020}. Notably, some studies leveraged lexical resources such as HowNet and WordNet to construct latent vectors or use them as guiding signals \citep{dong&dong2006,luo-etal-2018-leveraging,luo-etal-2018-incorporating,blevins-zettlemoyer-2020-moving,li2020explicit,scarlini2020sensembert,yang2020incorporating}. 

Recently, contextualized embeddings have been shown to capture essential aspects of lexical semantics (\citealp{Peters2018,Loureiro2019}) in the word sense disambiguation literature. Utilizing the sense vectors built from the contextual embeddings, \citet{Scarlini2020} found that a simple 1-nearest-neighbor algorithm achieves comparable performance with other supervised model architectures in the word sense disambiguation task. The result demonstrates that the encoded vectors could carry significant semantic information, which should be not only applicable when disambiguating the polysemous words, but beneficial to definition modeling. 

In the following, we propose a \texttt{Vec2Gloss} model to formulate the definition modeling as a sequence-to-sequence problem with an encoder-decoder architecture \citep{Mickus2019}. However, an important distinction is that the goal of Vec2Gloss is to decode the definition out of the encoded vectors while simultaneously tuning the encoder for an optimized semantic vector. Therefore, we leverage the pretrained mT5 \citep{Xue2021} text-to-text model architecture but impose a tight bottleneck between the encoder and decoder. Furthermore, as the decoder no longer has access to the full sentential context, the generated gloss could not depend on the collocations directly. Specifically, the decoder could only produce the gloss with the encoded semantic vectors and the learned gloss patterns.

\section{Vec2Gloss Model}

The goal of the Vec2Gloss model is to produce an intelligible gloss from a word's semantic vector based on Chinese Wordnet. The task is highly related to, although distinct from, the common NLP tasks. Specifically, the model's objective is more than obtaining an encoder representation and mapping a lexical word/sense into a vector. The vector must be optimized for decoding the gloss. On the other hand, the task is more than a standard autoregressive one, as the generated gloss must be conditioned on a vector rather than prompts or input sequences. An encoder-decoder architecture might be the closest option, but the standard task is to map between the input and output text. It is unclear what the model learns to decode the gloss from the semantic vector or just \emph{translate} it from the input text.

\begin{figure}
\centering
\includegraphics[width=0.9\linewidth]{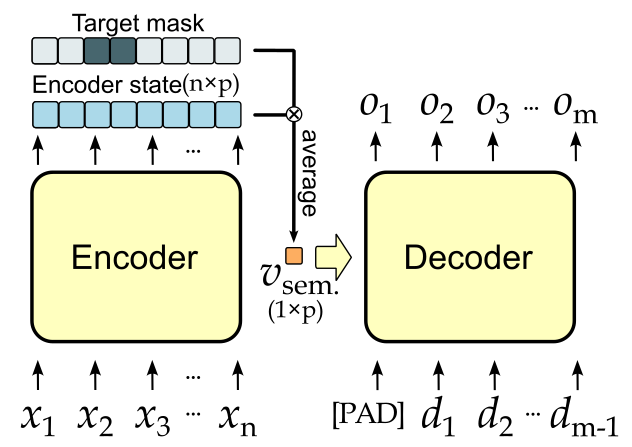}
\caption{The model architecture of Vec2Gloss. The model follows a general encoder-decoder architecture, but a bottleneck is imposed between the encoder and decoder. The decoder, instead of having access to the full encoder states, only \textit{sees} the target word's embeddings ($v_{sem}$).}
\label{fig.arch}
\end{figure}

To leverage the encoder-decoder architecture and simultaneously ensure the model relies on a semantic vector to decode the gloss, we impose a tight bottleneck between the encoder and decoder (Figure \ref{fig.arch}). The model's input is a sentence containing a target word. The input sentence is transformed by an encoder and results in a set of encoder states. Next, a predefined target mask is applied to the encoder states, and the target's encoder vectors are selected. These vectors are then averaged into a single vector and fed into the decoder, which learns to generate the gloss sequence. Notably, instead of mixing the encoder states with cross attentions in standard architecture, the decoder here only has one encoder vector to attend to. That is, the decoder cannot access the complete input sentences. Therefore, the encoder is driven to compress as much information as possible into the target word's semantic vector ($v_{sem.}$). On the other hand, the decoder must learn the gloss's regularities instead of relying on potential collocation cues between the word context and the gloss. Taken together, the model simultaneously learns the target word's semantic vector with the encoder, from which the decoder produces the gloss sequence.

To enhance the model's ability to capture the pattern of gloss sequences, we propose a denoising stage \emph{prior to} training for the vector-to-gloss task. In the denoising stage, a standard encoder-decoder architecture is used, and the model learns to reconstruct the corrupted spans in the glosses. The purpose is to pretrain the model to capture the regularities of gloss language better. Afterward, we impose the bottleneck between the encoder and decoder in the fine-tuning stage. The model takes as input a sentence containing a target word along with a target mask, and it needs to learn the target word's semantic vector with the encoder and generate as the output the gloss sentence entirely from the semantic vector.

\subsection{Denoising stage}

We first train the model with a denoising objective to better capture the patterns underlying the gloss language. Following the procedures of previous studies \citep{Lewis2020}, we prepare pairs of examples consisting of corrupted spans as inputs and the dropped-out spans as outputs. Such denoising objective is shown to perform well on downstream tasks and be computationally efficient as having shorter decoding sequences \citep{Raffel2020}. A pair of such examples is shown as follows, and the literal translations are in italics:

\vspace{1ex}
\noindent
{\small
\begin{tabular}{ll}
\textbf{Input} & 以文字媒介\sentinel{X}出來的訊息。 \\
               & \textit{using text medium} \sentinel{X}
                 \textit{-out information.} \\
\textbf{Target} & \sentinel{X}表達\sentinel{Y} \\
                & \sentinel{X}\textit{express}\sentinel{Y} \\
\end{tabular}
\vspace{1ex}
}

The \sentinel{X} and \sentinel{Y} stand for the special sentinel tokens; they are unique within an example. The spans are character-based and may not follow the word boundaries. The corrupted locations are randomly selected, and the lengths (number of characters) are randomly drawn from a Poisson distribution with $\lambda=2$, with the value clipped between 1 and 4 (inclusive). If the input sequence is longer than 20 characters, another corrupted span is created with the same parameter. The data are extracted from the word glosses of Chinese Wordnet. There are 26,118 pairs created for the denoising objective.

We use the pretrained T5 encoder-decoder architecture (\texttt{mt5-base}) to train the denoising objective \citep{Xue2021}. In the denoising stage, no bottleneck is applied. The model parameters are updated with AdamW optimizer. The learning rate is $10^{-4}$, $\beta_1$ and $\beta_2$ are 0.9 and 0.999, respectively, and weight decay is set to 0.01. A linear schedule is applied to the learning rate, and the batch size is set to 8. The model was trained for 3 epochs and took 30 minutes in an A5000 GPU. The trained model parameters are the starting point of the next fine-tuning stage.

\subsection{Fine-tuning stage}
The fine-tuning stage aims to learn the relationships between the target words embedded in the sentences and their glosses in CWN. In addition to the standard T5 encoder-decoder transformer-based architecture, a tight bottleneck is introduced between the encoder and decoder. Namely, only the target word's encoder states, which may consist of more than one token, are selected and averaged as the semantic vector, from which the decoder learns to produce a complete gloss sentence. 

The training data are extracted from the CWN's sense inventories. A training instance is created for each example sentence in a CWN sense. The instance is composed of a pair of input and target sequences. The input is an example sentence in which the target words are marked with a pair of angular brackets. The target sequences are the glosses preceded by the part-of-speeches of the senses and followed by a Chinese full-width period. There are 76,969 instances in the training dataset and 8,553 pairs in the evaluation dataset. A sample instance is shown as follows:

\vspace{2
ex}
\noindent
{\small
\begin{tabular}{ll}
\textbf{Inp.} & 她不知道為了什麼事而默默不\textlangle{}語\textrangle{}。 \\
               & \textit{She didn't} \sentinel{say} \textit{a word for some reason.}\\
\textbf{Tgt.} & VA。透過發聲器官，用語音傳送訊息。 \\
                & \textit{VA. Using vocal organs to convey} \\
                & \textit{a message with speech.}\\
\end{tabular}
\vspace{1ex}
}

The model architecture closely follows the standard T5; thus, the trained weights in the denoising stage are directly applicable to this model. Notably, the target words' angular brackets are preprocessed and removed to produce the target mask. The mask is used to select the relevant encoder states and generate the semantic vector for the decoder. Therefore, the cross attention of the decoder will always receive a single vector as input. In training time, the model is trained as a text-to-text task. However, in inference time, the encoder and decoder could work independently. That is, the encoder could be used to obtain a semantic vector from a given sentence. The semantic vector can then be flexibly transformed before sending it into the decoder for gloss generation.


The training procedure is the same as the previous stage, the only difference being that the epoch number is 10 for this stage. The training time is 100 minutes in an A5000 GPU.

\subsection{Automatic evaluations}
The automatic evaluation of the definition generation is shown in Table ~\ref{tab.metrics}, which presents the BLEU and METEOR scores for each lexical category. The overall score is .41 for BLEU and .62 for METEOR. It is noteworthy that the lowest category is the noun (N), while the highest one is the proper name (Nb). The higher score of the proper name may be attributed to words used as a family or foreign names in CWN. Their definitions are short and are thus more likely to be captured by the model. These names account for 188 items in the proper names category. On the other hand, it is less clear how to interpret the automatic metrics in other categories. The scores only indicate the textual difference between the generated and the reference gloss. At a given score level, the generated gloss might be unintelligible to the human reader or just a paraphrase of the reference gloss. Therefore, we study the generated glosses further with human evaluations, including a rating experiment, a gloss dependency analysis, and a token dependency analysis.

\begin{table}
\centering
\begin{tabular}{rrrr}
\toprule
POS  &   N   &    BLEU      &  METEOR \\
\midrule
N    &  2,801  & .35(.01) & .59(.01) \\
V    &  4,376  & .43(.01) & .63(.01) \\
D    &    432  & .41(.02) & .62(.02) \\
O    &    530  & .41(.02) & .63(.01) \\
Nb   &    414  & .63(.02) & .74(.02) \\
\midrule
All  &  8,553  & .41(.01) & .62(.01) \\
\bottomrule
\end{tabular}
\caption{Automatic evaluation metrics on different lexical categories, which are nouns (N), verbs (V), adverbs (D), others (O), and proper names (Nb). Numbers in parentheses are standard errors.}
\label{tab.metrics}
\end{table}

\section{Human Evaluations}

\subsection{Rating experiment}
In the experiment, we resort to human raters to evaluate the capability of the generated definitions, i.e., their semantic interpretability and syntactic well-formedness. The task is designed as a multiple-choice task with only one correct answer. A total of 140 entries, each consisting of a definition in Chinese and a list of four-word options, are provided. Materials used for the experiment are derived from two sources: Academia Sinica Balanced Corpus of Modern Chinese (henceforth ASBC, \citealp{huang1998}) and CWN.

Among the 140 test items, 40 are new words with their definitions derived from our Vec2Gloss model (namely \emph{V2G:ex vivo}). We extract words only composed of Chinese characters, remove proper nouns, and filter out words occurring less than 10 times in the corpus. The target words, i.e., the correct answers, are randomly and equally selected from four different lexical categories: nouns, verbs, adverbs, and others. The incorrect options of each question are of the same word class, randomly selected from the same collection of words derived from ASBC. The remaining 100 words are all taken from the evaluation dataset, with proper names being excluded. Among the 100 words, 20 use definitions from CWN, 80 from model generation (namely \emph{V2G:in vivo}). The word class composition is the same for the ones from CWN, and the target words are randomly selected from the dataset and equally divided according to the word class.

Five native Chinese speakers majoring in linguistics were recruited as raters. Their first task was to determine the most suitable term from a set of four options based on its given definition. Next, the raters were asked to evaluate the semantic interpretability of a definition using a five-point acceptability judgment scale. That is, they were required to rate on a scale of one to five, to what extent the definition can well explain the word that had been chosen as the correct answer from the previous task. Finally, the raters were asked to evaluate the syntactic well-formedness of a definition, i.e., how much the rater accepted the definition as being well-formed based on his or her internal grammar, on a five-point acceptability judgment scale. Table~\ref{tab.stat.all} shows the evaluation results with these two metrics, where we can see the proposed Vec2Gloss model achieves decent performance in comparison with the original glosses in CWN.

\begin{table}[b]
\centering
\resizebox{\columnwidth}{!}{
\begin{tabular}{rrrr}
\toprule
Source & Correctness & Mean{\textsubscript{sem}} & Mean{\textsubscript{syn}}\\
\midrule
CWN & .95(.02) & 4.47(.15) & 4.82(.10) \\
V2G:in vivo & .88(.03) & 3.51(.16) & 4.58(.09) \\
V2G:ex vivo & .86(.04) & 2.53(.22) & 4.51(.12) \\
\bottomrule
\end{tabular}
}
\caption{Human evaluation results for definitions generated from different sources, with Mean{\textsubscript{sem}} and Mean{\textsubscript{syn}} representing the mean value of semantic interpretability and syntactic well-formedness, respectively.}
\label{tab.stat.all}
\end{table}

More detailed results for evaluations of vector-generated glosses are shown in Table \ref{tab.stat.pos}. While the mean values for syntactic well-formedness for both \emph{V2G:in vivo} and \emph{V2G:ex vivo} are considerably high across all the 4 lexical categories, the semantic interpretability for \emph{V2G:ex vivo} scores less than for \emph{V2G:in vivo}. In spite of the inferior interpretability, the multiple-choice task for \emph{V2G:ex vivo} still achieves over 80\% correct rates in every category, similar to the results of \emph{V2G:in vivo}. Moreover, the semantic scores of nouns are lower than other categories for both sources. To further investigate possible reasons that account for the results, we conduct a gloss dependency analysis.

\begin{table*}
\centering
\begin{tabular}{rrrrrrr}
\toprule
\multirow{2}{*}[-0.3em]{\textbf{POS}} &
\multicolumn{3}{c}{V2G:in vivo} &
\multicolumn{3}{c}{V2G:ex vivo}\\\cmidrule(lr){2-4} \cmidrule(lr){5-7}
& Correctness & Mean{\textsubscript{sem}} & Mean{\textsubscript{syn}} & Correctness & Mean{\textsubscript{sem}} & Mean{\textsubscript{syn}}\\
\midrule
N & .94 (.04) & 3.18 (.35) & 4.14 (.25) &
  .86 (.08) & 1.92 (.40) & 4.32 (.34) \\
V & .89 (.06) & 3.63 (.34) & 4.79 (.10) & 
  .86 (.08) & 2.74 (.46) & 4.48 (.27) \\
D & .84 (.06) & 3.75 (.31) & 4.69 (.18) &
  .84 (.07) & 2.76 (.43) & 4.74 (.16) \\
O & .85 (.06) & 3.47 (.32) & 4.70 (.16) & 
  .86 (.10) & 2.70 (.45) & 4.50 (.20) \\
\bottomrule
\end{tabular}
\caption{Human evaluation results for different lexical categories of definitions generated from \emph{V2G:in vivo} and \emph{V2G:ex vivo}. The semantic evaluation scores of noun are lower than other categories for both sources.}
\label{tab.stat.pos}
\end{table*}

\subsection{Gloss dependency analysis}
\label{sec.gloss.analysis}

Two indices are computed for each token to represent their reliance on the preceding contexts and the semantic vector, respectively. First, the token likelihood under the full context and the original semantic vector ($p_{\textrm{full}}$) is compared to the one with all of its preceding contexts masked when decoding. If a token is mostly determined by the context alone, the context masking would significantly impact the token likelihood ($p_{\textrm{mask}}$). Hence, the negative likelihood ratio ($\delta_{\textrm{sem}}$) will be larger. Similarly, if a token is primarily driven by the semantic vector, replacing it while leaving the preceding context intact will lower the likelihood ($p_{\textrm{rep}}$) and make the ratio ($\delta_{\textrm{ctx.}}$) larger. Specifically, the semantic vector ($v_{sem.}$ coming from the encoder) is replaced with another word's semantic vector from the same lexical category. The indices are all calculated using the shifted reference glosses of each sense as the decoder inputs.

\vspace{-0.5cm}
\begin{align}
\delta_{\textrm{sem}} & = -\log(p_{\textrm{rep}} / p_{\textrm{full}}) \nonumber \\
\delta_{\textrm{ctx}} & = -\log(p_{\textrm{mask}} / p_{\textrm{full}}) \nonumber
\end{align}

\begin{figure*}
\centering
\includegraphics[width=0.9\textwidth]{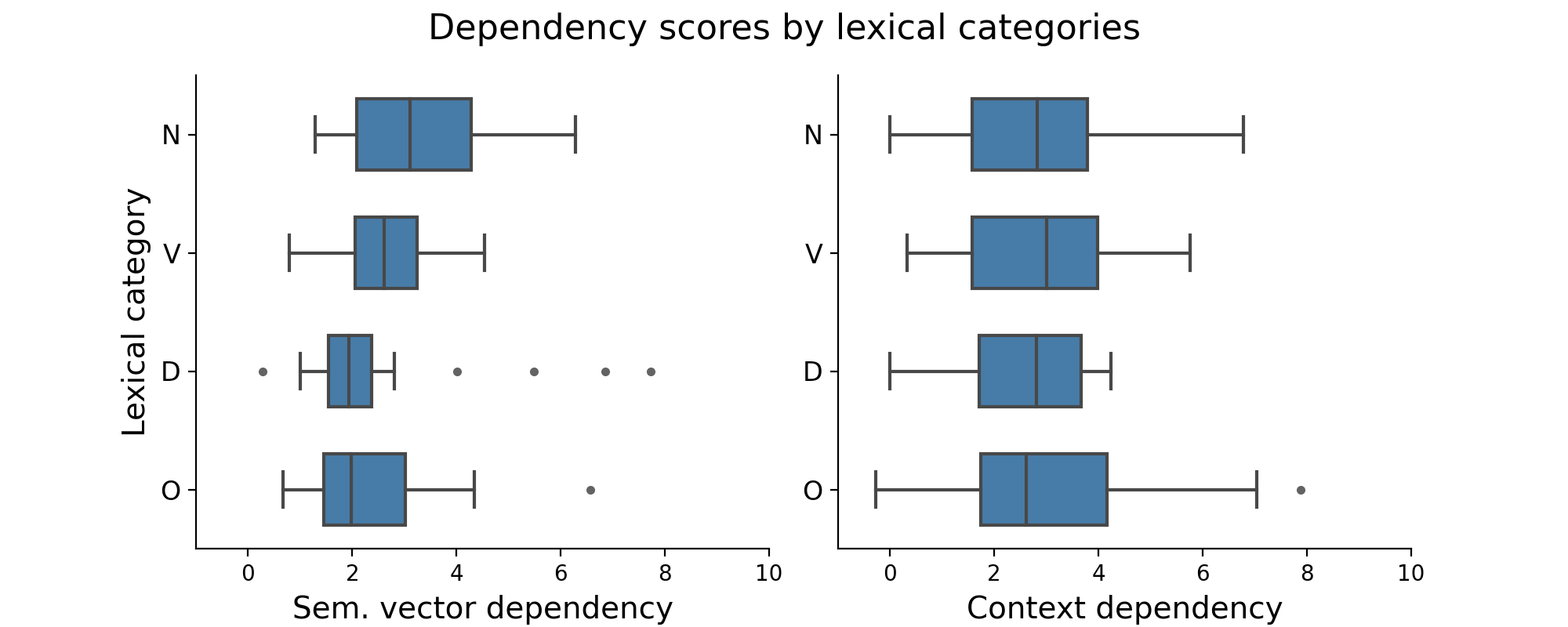}
\caption{Dependency scores by each lexical category. The left panel shows the semantic dependency and the right one shows the context dependency scores. The letters along the vertical axis denote the lexical categories: nouns (N), verbs (V), adverbs (D), and others (O).}
\label{fig.deps}
\end{figure*}

Each token's indices, $\delta_{\textrm{sem}}$ and $\delta_{\textrm{ctx.}}$, are averaged to produce the gloss-level indices. The results are shown in Figure \ref{fig.deps}. We first observe that while the contextual dependency is comparable across four different lexical categories, the semantic vector dependency indices show more differences. Specifically, the nouns' glosses have higher semantic dependency scores, followed by verbs, adverbs, and others. These results echo the human ratings, in which the syntactic ratings are similar in all categories, while nouns are significantly worse in semantic rating scores. The difference in semantic dependency may indicate that nouns are more likely to be used as nominal predicates, and they \emph{categorize} referents in a class with a holistic set of properties. In contrast, the adverbs, which have the lowest semantic dependency score here, only \emph{describe} things by adding a single property to the characterization of the referent \citep{Baker2017,Bolinger1980}. However, if this is the case, the adverbs present interesting cases to study further. Although they are relatively unaffected by semantic vector manipulation, they still carry semantic meaning such as manner, mean, or instrument into their scopes \citep{Lyons1977,Lakoff1968}. Therefore, we should observe some tokens are more pertinent to the semantic vector than others in a more fine-grained analysis.

\subsection{Token dependency analysis}
\label{sec.tok.dep}
Following that, we manually identify chunks (\emph{semantic constituency}) in the gloss and annotate their semantic types. The \emph{chunk} as defined here is a \emph{significant element} which functions as a semantic type-carrying unit (cf. \citealp{gerdes2013defining}). 
We selected 244 adverbs from CWN, whose gloss contains the word ``\textit{事件}'' \textit{shjiàn} `event', as they describe an explicit event structure in the glosses. Each gloss was first segmented into length-variant chunks and manually tagged with its corresponding semantic type. The gloss of the first words is not annotated with semantic types, as they regularly follow a gloss pattern based on their lexical category. For example, the glosses of adverbs start with the word \textit{表 biǎo} `indicate,' such as the following example (the gloss of \textit{接連 jiēlián} `in a row'). 

\vspace{1ex}
\noindent
{\small
\begin{tabular}{ll}
\textbf{Gloss} & 表/同一事件/在/後述時段/中/持續/發生。 \\
             & \textit{To express the same event continuously} \\
                & \textit{happens during the later-mentioned period.} \\
\textbf{Annot.} & -{}-/Event/Preposition/Time/Preposition/\\
                & Modifier/Action\\
\end{tabular}
\vspace{1ex}
}

There are 905 chunks and 19 unique semantic types annotated in this dataset. Six semantic types ({\tt event, action, modifier, pre/post-position, negation, others}) occurring at least 25 times (10\% of the glosses count) are selected for further analysis, which accounts for 59\% of annotated chunks. Figures in the Appendix show their distribution in different positions in the gloss.


\begin{figure}
\centering
\includegraphics[width=\linewidth]{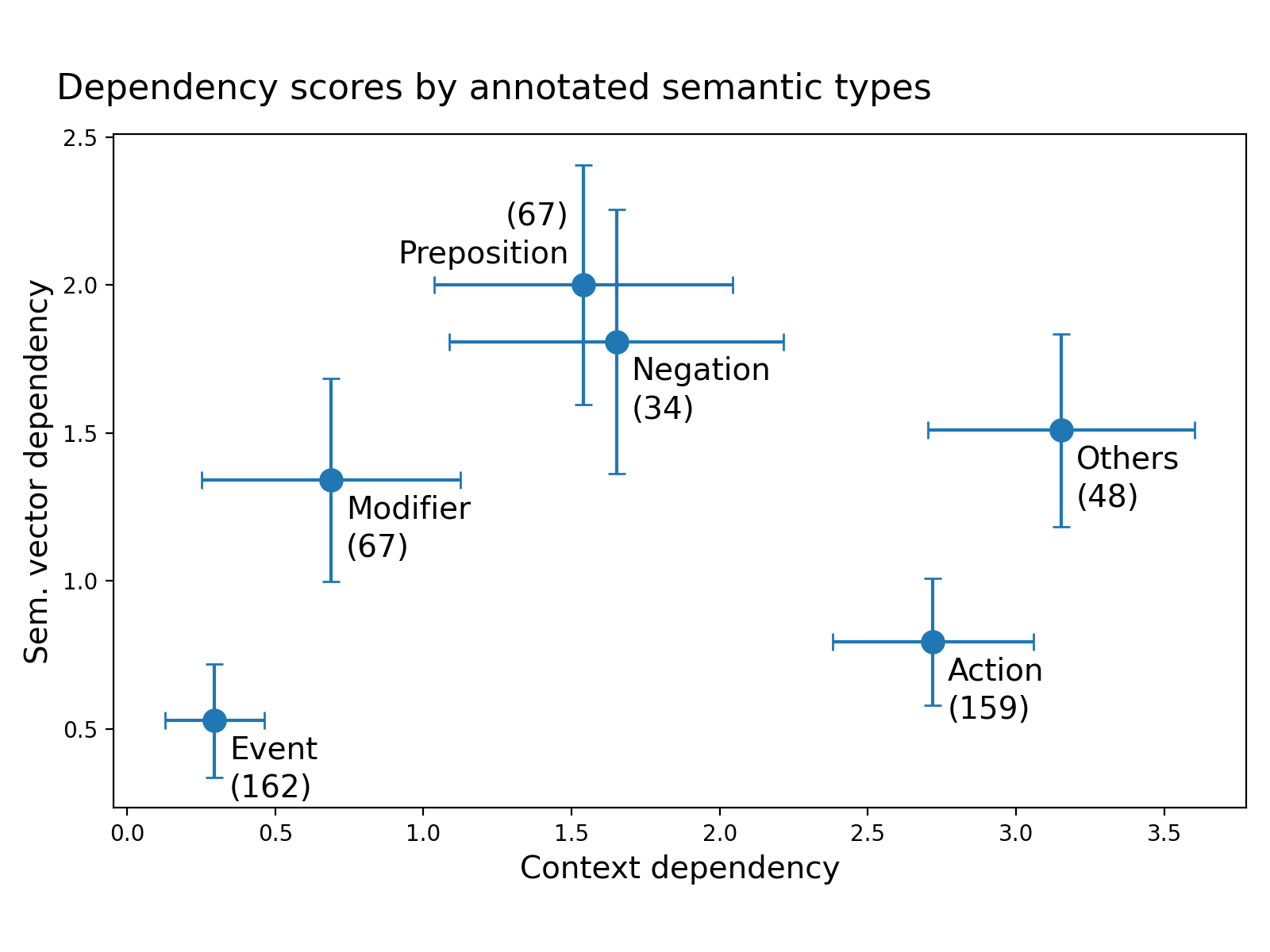}
\caption{The dependency scores of six annotated semantic types. The error bars denote one standard error of semantic or context dependency scores. Numbers in parentheses are the member count of the type.}
\label{fig.chunk}
\end{figure}

The token level indices are computed as in Sec.~\ref{sec.gloss.analysis}. Notably, since the annotated glosses may have multiple example sentences in CWN, we extract and average the semantic vectors from each sentence to represent the target words. The context and semantic vector dependency scores are computed for each token and averaged by their semantic types. The results are shown in Figure \ref{fig.chunk}.

In alignment with the observation in gloss-level analysis, there are distinctive dependency patterns across different semantic types. In particular, the Action categories are higher in contextual dependency but relatively low in semantic ones. While Action words are usually the main verbs in the glosses, the type distribution is highly skewed. The first three common action words, \textit{發生 fāshēng} `occur', \textit{做 zuò} `make' , \textit{進行 jìnxíng} `undergo' already account for 50\% of all action words. Thus, the contextual dependency scores may reflect the constrained word usage given the adverb gloss. In contrast, the Preposition and Negation types are relatively high in semantic vector dependency. It might result from the fact that prepositions are used to introduce the relating complements, and the decoder needs guidance from the semantic vectors to select the exact relations for the gloss. Similarly, the negation words are hard to capture by syntagmatic relations from the context \citep{Aina2019,Ettinger2020}; thus, the decoder has to rely on additional cues from the semantic vectors. Consistently, words are highly predictable given the adverb glosses, i.e., the Event type, are low on both dependency scores. However, it is interesting for future work to analyze all the patterns or semantic types of the glosses and investigate the exact roles of contextual and semantic dependencies in the gloss patterns.

\section{Conclusion}
This paper introduces a definition model called \emph{Vec2Gloss}. In addition to training a gloss generation model that directly decodes the gloss from one semantic vector while optimizing the encoder simultaneously, we also examine the generated glosses to gain a deeper understanding of the lexical-semantic information captured by the model. The systematic study of glosses is made possible by the systematic gloss patterns provided by CWN. In the experiment, we observe that nouns and verbs are more complicated in terms of their higher semantic dependency, while adverbs are less so. Furthermore, in the token-level analysis, we also found models need additional guidance from semantic vectors for negation words and selecting proper prepositions.
Understanding semantic vectors in high-dimensional space is challenging; the reformulated definition modeling task provides another way to look into the distributed semantic vectors. Future research can focus on how systematic gloss patterns help us further understand the intricate lexical categories, such as nouns and verbs, and determine the optimal gloss patterns through the application of distributed semantic models.


\bibliography{custom}
\bibliographystyle{acl_natbib}

\appendix
\section{Appendix}
\label{sec:appendix}
\begin{figure*}
\centering
\includegraphics[width=0.8\textwidth]{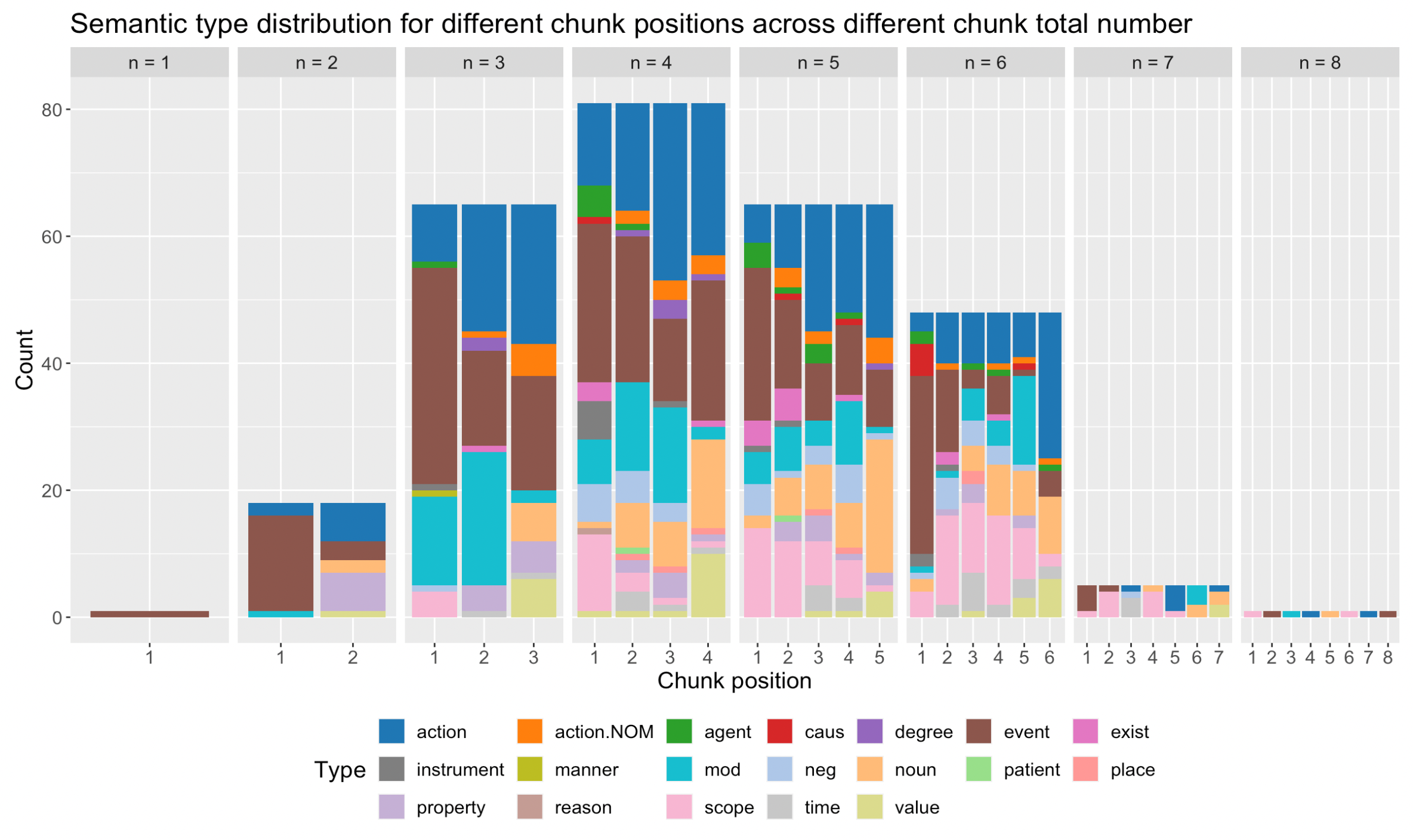}
\caption{Distribution of chunk frequencies of all semantic types by positions and sequence lengths. }
\label{fig.deps.chunk1}
\end{figure*}

\begin{figure*}
\centering
\includegraphics[width=0.8\textwidth]{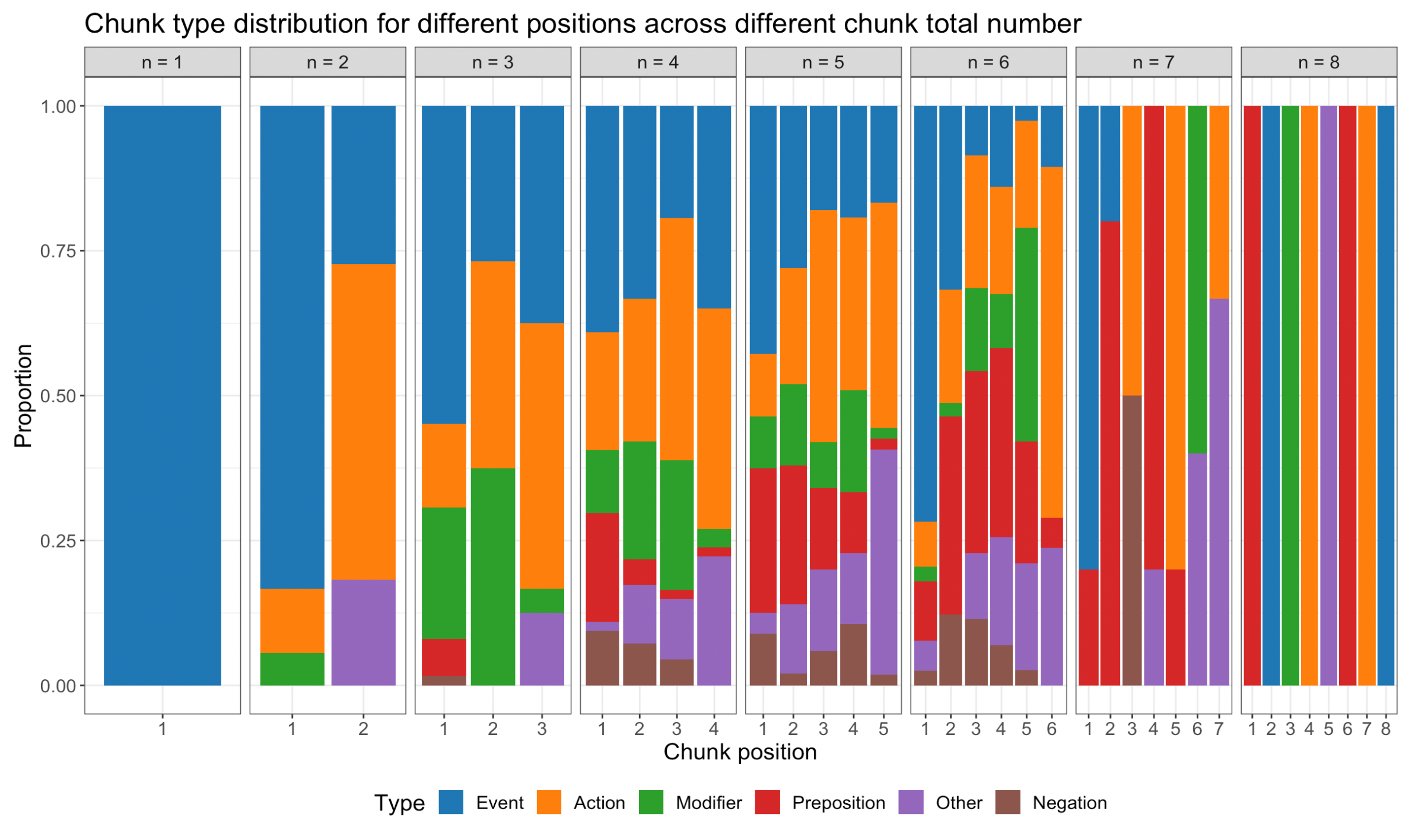}
\caption{Proportion of chunk types by positions and sequence lengths of semantic types occurs in more than 10\% of sequences.}
\label{fig.deps.chunk2}
\end{figure*}

Table \ref{tab.gensamples} illustrates some examples of the model-generated glosses. Figure \ref{fig.deps.chunk1} and Figure \ref{fig.deps.chunk2} shows the statistics of semantic type annotations in Section \ref{sec.tok.dep}.

\begin{table*}
\small
\centering
\begin{tabular}{rp{6.5cm}p{7.5cm}}
\toprule
  &  Input  & Generated \\
\midrule
1 & 他\sentinel{還}沒開口。 \newline \textit{He hasn't \sentinel{yet} spoken.}
  & Dfa。表事情尚未完成。 \newline \textit{Dfa. Describing the situation not having finished} \\
2 & 他還沒\sentinel{開}口。 \newline \textit{He hasn't yet \sentinel{spoken}.}
  & VC。比喻提出要求。 \newline 
    \textit{VC. Making a request. }\\
3 & 我\sentinel{開}了一個會。 \newline \textit{I \sentinel{had} a meeting.} 
  & VC。進行會議。 \newline \textit{VC. Holding a meeting. } \\
4 & 這\sentinel{彰顯}出重要的價值。 \newline \textit{This \sentinel{exemplifies} an important value.} 
  & VJ。顯現出後述事物或特質。\newline 
    \textit{VJ. Showing the quality of the following situation.} \\
\bottomrule
\end{tabular}
\caption{Examples of the model-generated glosses. The first three instances include the target words already existing in CWN, but the sentences are all new to the model. The second and third ones show the context dependencies of the generated glosses. The target word of the last instance is also new to the model, and the model still generates a plausible gloss. Dfa. Degree adverb. VC. Action transitive verb. VJ. Stative transitive verb.}
\label{tab.gensamples}
\end{table*}

\end{CJK*}
\end{document}